%% file: main.tex
\documentclass[10pt,twocolumn,letterpaper]{article}

\usepackage{wacv}
\usepackage{times}
\usepackage{epsfig}
\usepackage{graphicx}
\usepackage{amsmath}
\usepackage{amssymb}
\usepackage{booktabs}
\usepackage{caption}
\usepackage{multirow}
\usepackage{algorithm2e}

\def\wacvPaperID{1169} 

\wacvalgorithmstrack   

\wacvfinalcopy 

\ifwacvfinal
\usepackage[breaklinks=true,bookmarks=false]{hyperref}
\else
\usepackage[pagebackref=true,breaklinks=true,colorlinks,bookmarks=false]{hyperref}
\fi

\begin{document}

\title{More Control for Free! Image Synthesis with Semantic Diffusion Guidance}

\author{Xihui Liu$^{1,4}\thanks{This work was done when Xihui Liu was a postdoc at UC Berkeley.}$~~~~~~~
Dong Huk Park$^1$~~~~~~~
Samaneh Azadi$^1$~~~~~~~
Gong Zhang$^{2,3}$\\   
Arman Chopikyan$^2$~~~~~
Yuxiao Hu$^2$~~~~~
Humphrey Shi$^{2,3}$~~~~~
Anna Rohrbach$^1$~~~~~
Trevor Darrell$^1$\\
$^1$UC Berkeley~~~~~$^2$Picsart AI Research (PAIR)~~~~~$^3$University of Oregon~~~~$^4$The University of Hong Kong}

\makeatletter
\let\@oldmaketitle\@maketitle
\renewcommand{\@maketitle}{\@oldmaketitle
\vspace{-5pt}
\centering
\includegraphics[width=0.95\linewidth]{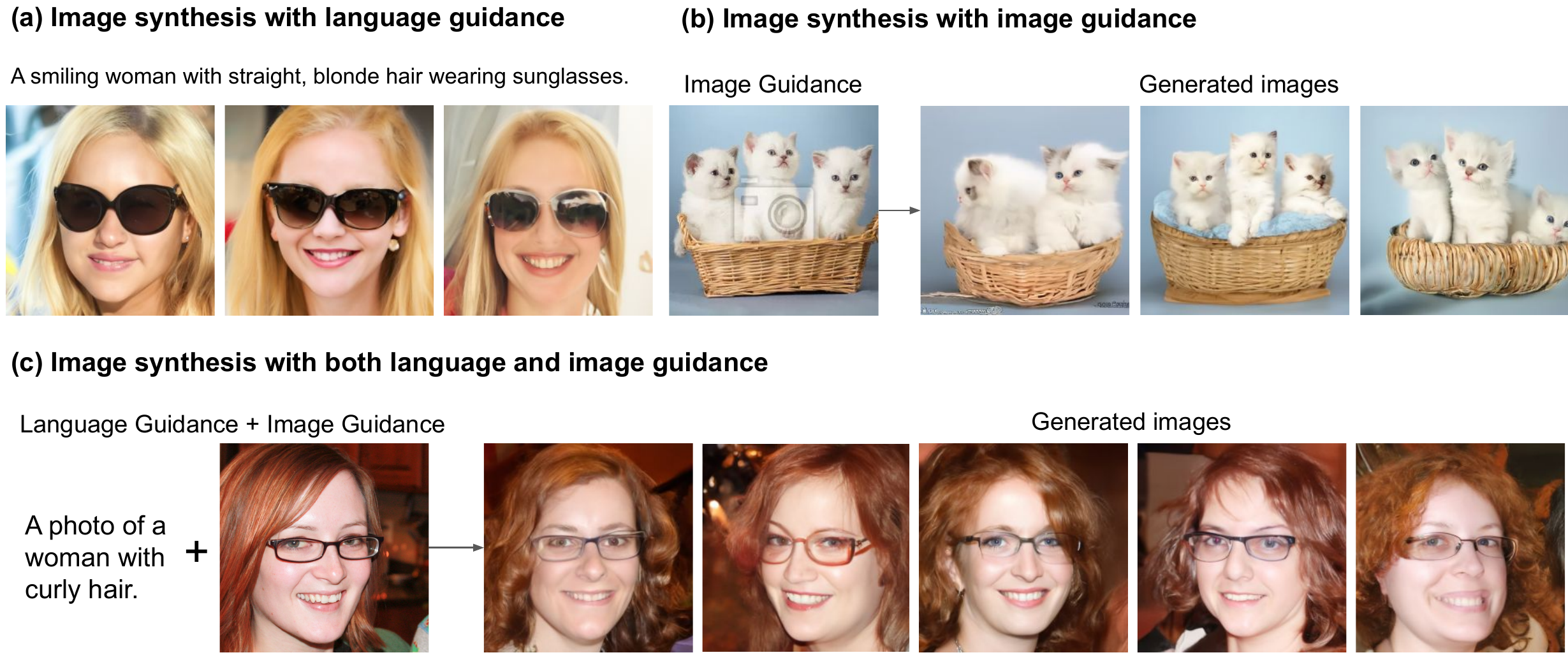}
\captionof{figure}{We incorporate flexible semantic guidance into diffusion models for image synthesis. Our method allows fine-grained control via language, image, or multimodal guidance. It can be applied to datasets without paired image-text data.}
\vspace{-4mm}
\label{fig:intro}
\bigskip\bigskip}
\makeatother

\maketitle

\begin{abstract}
Controllable image synthesis models allow creation of diverse images based on text instructions or guidance from a reference image. Recently, denoising diffusion probabilistic models have been shown to generate more realistic imagery than prior methods, and have been successfully demonstrated in unconditional and class-conditional settings. We investigate fine-grained, continuous control of this model class, and introduce a novel unified framework for semantic diffusion guidance, which allows either language or image guidance, or both. Guidance is injected into a pretrained unconditional diffusion model using the gradient of image-text or image matching scores, without re-training the diffusion model.  We explore CLIP-based language guidance as well as both content and style-based image guidance in a unified framework. Our text-guided synthesis approach can be applied to datasets without associated text annotations. We conduct experiments on FFHQ and LSUN datasets, and show results on fine-grained text-guided image synthesis, synthesis of images related to a style or content reference image, and examples with both textual and image guidance.
\footnote{Project page \url{https://xh-liu.github.io/sdg/}}
\end{abstract}

\input{sections/intro}
\input{sections/related}

\input{sections/method}

\input{sections/experiments}

\input{sections/conclusion}

{\small
\bibliographystyle{ieee_fullname}
\bibliography{egbib}
}

\end{document}

%% file: sections/intro.tex
\vspace{-3mm}
\section{Introduction}\label{sec:intro}

Image synthesis has made great progress in recent years~\cite{karras2019style,brock2018large,park2019semantic,ramesh2021zero,dhariwal2021diffusion}. In addition to the goal of generating high-quality photo-realistic images, fine-grained control over the generated images is also an important desideratum when assisting users with art creation and design.

Previous works have explored controllable image synthesis by adding different conditions, including language~\cite{zhang2017stackgan,xu2017attngan,ramesh2021zero}, attributes~\cite{zhu2020domain,shoshan2021gan,xiao2018elegant}, scene graphs~\cite{dhamo2020semantic}, and user sketches or scribbles~\cite{bau2019semantic}. Specifically, text-to-image synthesis, as shown in Figure~\ref{fig:intro}-(a), aims to generate images based on text instructions, by adding text embeddings as conditional information to the image generation network. However, most previous text-to-image synthesis methods require image-caption pairs for training, and cannot generalize to datasets without text annotations.

Besides text instructions, users may want to guide the image generation model with a reference image. E.g., a user might want to generate cat images which look similar to a given photo of a cat in terms of its appearance. This information cannot be easily described by language, but can be provided via a reference image, as shown in Figure~\ref{fig:intro}-(b). Furthermore, a user may want to provide both language and image guidance. For example, a user might seek to generate ``a woman with curly hair'' that looks similar to a reference image of a woman with red hair, as shown in Figure~\ref{fig:intro}-(c).

Current image-conditioned synthesis techniques either only transfer the ``style'' of a reference image to a target image~\cite{huang2018multimodal,azadi2018multi} or are restricted to the domains with well-defined structure, e.g. human or animal faces~\cite{huang2018multimodal,xiao2018elegant}. They cannot generate diverse images with various poses, structures and layouts based on a single reference image.

We propose \emph{Semantic Diffusion Guidance (SDG)}, a unified framework for text-guided and image-guided synthesis that overcomes these limitations. Our model is based on 
denoising diffusion probabilistic models (DDPM)~\cite{ho2020denoising} which generate an image from a noise map and iteratively removes noise to approach the data distribution of natural images. 

We inject the semantic input by using a guidance function to guide the sampling process of an unconditional diffusion model. This enables more controllable generation in diffusion models and gives us a unified formulation for both language and image guidance. Specifically, our language guidance is based on the image-text matching score predicted by  CLIP~\cite{radford2021learning} finetuned on noised images. As for the image guidance, depending on what information we seek in the image, we define two options: content and style guidance. The flexibility of the guidance module allows us to inject either language or image guidance alone or both at once into any unconditional diffusion model without the need for re-training. We propose a self-supervised scheme to finetune the CLIP image encoder without text annotations, from which we obtain the guidance model with minimal cost.

Our unified framework is flexible and allows fine-grained semantic control in image synthesis as shown in Figure~\ref{fig:intro}. We show that our model can handle: (1) Text-guided image synthesis with fine-grained text queries on any dataset \emph{without language annotations}; (2) Image-guided image synthesis with content or style control from the input image, which generates diverse images with different pose, structure, and layout; (3) Multi-modal guidance for image synthesis with both language and image input. Our flexible guidance network can be injected into off-the-shelf unconditional diffusion models, \emph{without the need for re-training} the diffusion model. We further present a self-supervised efficient finetuning scheme for the CLIP guidance model which does not require textual annotations. We conduct experiments on FFHQ~\cite{gal2021stylegan} and LSUN~\cite{yu2015lsun} datasets to validate the quality, diversity, and controllability of our generated images, and show various applications of our proposed Semantic Diffusion Guidance.

%% file: sections/related.tex
\section{Related Work}\label{sec:related}

\textbf{Text-guided Synthesis} 
Pioneered by GAN-INT-CLS~\cite{reed2016generative} and GAWWN~\cite{reed2016learning}, conditional generative adversarial networks (GANs)~\cite{goodfellow2014generative} have been the dominant framework for text-based image synthesis. Various methods have been studied
leading to significant improvements in editing quality and correctness~\cite{xu2018attngan,zhu2019dm,tao2020df,li2019controllable,dong2017semantic,zhu2017your,mao2019bilinear,nam2018text,gunel2018language,li2020manigan,chen2018language,joseph2019c4synth,qiao2019mirrorgan}.
Recent work DALL-E~\cite{ramesh2021zero}
shows promising results with transformers~\cite{vaswani2017attention} and discrete VAE~\cite{razavi2019generating} by leveraging web-scale data. A concurrent work GLIDE~\cite{nichol2021glide} adapts classifier-free guidance for large diffusion models and large-scale training for text-guided image synthesis.
Despite great advancements, prior methods require paired image-text annotations which limits the application to certain datasets or requires large amount of data and computational resources for training. 
Our proposed framework is able to generate images on multiple domains given detailed text prompts, requiring neither image-text paired data from those domains nor large amount of compute to train the text-guided image synthesis model.

\textbf{Image-guided Synthesis} 
Image-guided synthesis aims to generate diverse images with the constraint that they all should resemble a given reference image in terms of content or style.
Many style transfer works fall under this category where the content of the input image must be preserved while the style of the reference image is transferred~\cite{gatys2016image,li2016combining,gu2018arbitrary,gatys2017controlling,kolkin2019style,huang2017arbitrary,li2017universal,li2019learning,park2020swapping,abdal2019image2stylegan}, yet they struggle to generate diverse images.
Some works study image synthesis guided by the content of the reference images. 
ILVR~\cite{choi2021ilvr} proposes a way to iteratively inject image guidance to a diffusion model,
yet it exhibits limited structural diversity of the generated images.
Instance-Conditioned GAN~\cite{casanova2021instance} uses nearest neighbor images of a given reference for adversarial training  
to generate structurally diverse yet semantically relevant images. Yet, it requires training the GAN model with instance-conditioned techniques.
Our approach demonstrates better controllability as different types of image guidance are proposed where users can decide how much semantic, structural, or style information to preserve by using different types and scales of guidance, while not requiring to re-train the unconditional diffusion model.

\textbf{Diffusion Models} 
Diffusion models are a new type of generative models consisting of a forward process (signal to noise) and a reverse process (noise to signal). 
The denoising diffusion probabilistic model (DDPM)~\cite{sohl2015deep,ho2020denoising} is a latent variable model where a denoising autoencoder
gradually transforms Gaussian noise into signal.
Score-based generative model~\cite{song2019generative,song2020improved,song2020score} trains a neural network to predict the score function 
which is used to draw samples via Langevin Dynamics. 
Diffusion models have demonstrated comparable or superior image quality compared to GANs while exhibiting better mode coverage and training stability.
They have also been explored for conditional generation such as class-conditional generation, image-guided synthesis, and super-resolution
~\cite{song2020score,dhariwal2021diffusion,choi2021ilvr,meng2021sdedit}. 
Concurrent work~\cite{avrahami2021blended} explored text-guided image editing with diffusion models.
Dhariwal~\etal~\cite{dhariwal2021diffusion} proposed classifier guidance for class-conditional image synthesis with diffusion models.
Based on the guidance algorithm proposed in~\cite{dhariwal2021diffusion}, we further explore whether diffusion models can be semantically guided by text or image, or both to synthesize images.

\textbf{CLIP-guided Generation and Manipulation} 
CLIP~\cite{radford2021learning} is a powerful vision-language joint embedding model trained on large-scale images and texts. Its representations have been shown to be robust and general enough to perform zero-shot classification and various vision-language tasks on diverse datasets. 
StyleCLIP~\cite{patashnik2021styleclip} and StyleGAN-NADA~\cite{gal2021stylegan} have demonstrated that CLIP enables text-guided \textit{image manipulation} and \textit{domain adaptation of image generation} without domain-specific image-text pairs. DiffusionCLIP~\cite{kim2021diffusionclip} uses CLIP for language-based image editing, and Blended Diffusion~\cite{avrahami2021blended} explores mask-guided image editing with CLIP.
However, the application to \textit{image synthesis} has not been explored.
Our work investigates text and/or image guided synthesis using CLIP and unconditional DDPM.

%% file: sections/method.tex
\vspace{-3mm}
\section{Semantic Diffusion Guidance}\label{sec:approach}

We propose \textit{Semantic Diffusion Guidance (SDG)}, a unified framework that incorporates different forms of guidance into a pretrained unconditional diffusion model. SDG can leverage language guidance, image guidance, and both, enabling controllable image synthesis. The guidance module can be injected into any off-the-shelf unconditional diffusion model without re-training or finetuning it. We only need to finetune the guidance network, which is a CLIP~\cite{radford2021learning} model in our implementation, on the images with different levels of noise. We propose a self-supervised finetuning scheme, which is efficient and does not require paired language data to finetune the CLIP image encoder.

In Section~\ref{subsec:guiding}, we review the preliminaries on diffusion models, and introduce our approach for injecting guidance 
for controllable image synthesis. In Section~\ref{subsec:language}, we describe the language guidance which enables the unconditional diffusion model to perform text-to-image synthesis. In Section~\ref{subsec:image}, we propose two types of image guidance, which take the content and style information from the reference image as the guidance signal, respectively. In Section~\ref{subsec:clip}, we explain how we finetune the CLIP network without requiring text annotations in the target domain.

\begin{figure}[t]
\begin{center}
\includegraphics[width=0.95\linewidth]{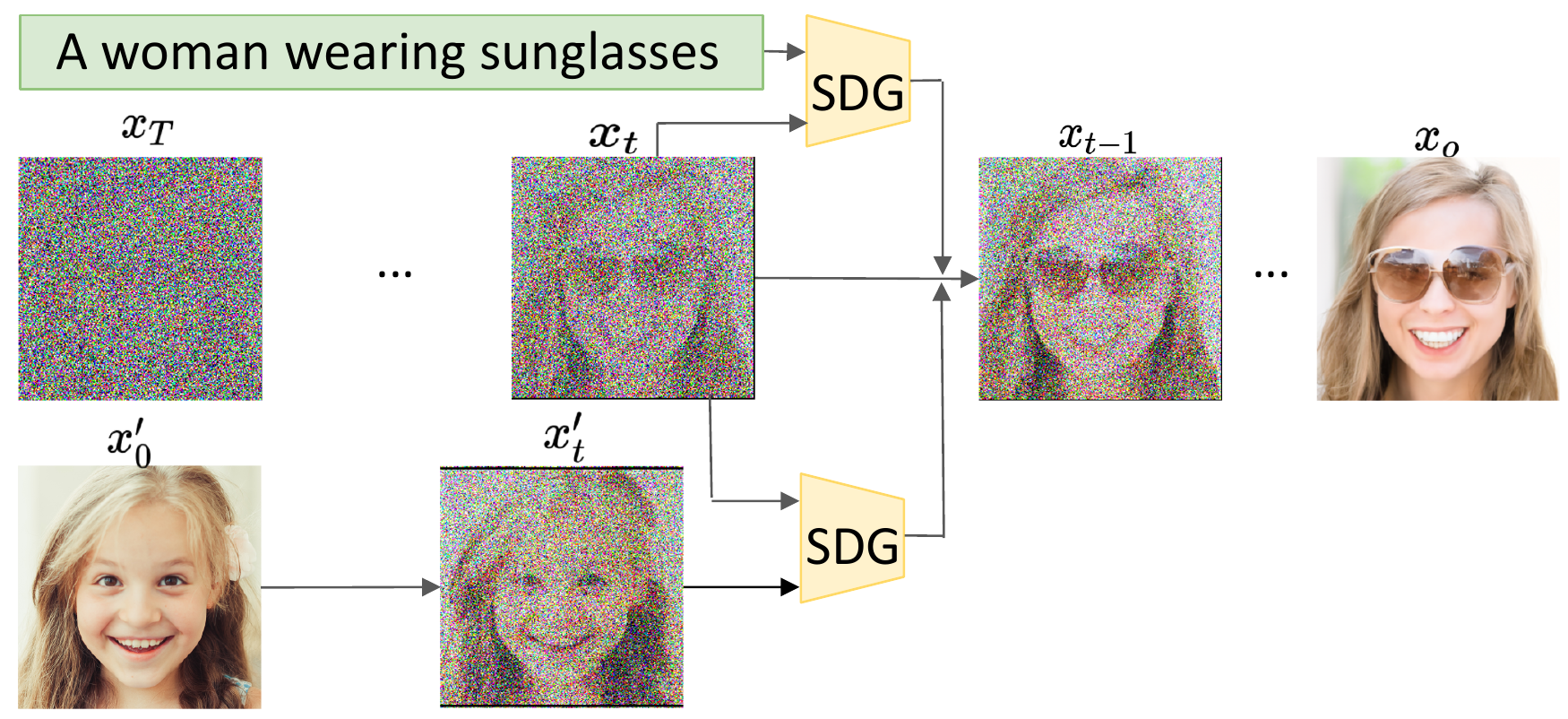}
\vspace{-2mm}
\caption{
Our method is based on the DDPM model which generates an image from a noise map by iteratively removing noise at each timestep. 
We control the diffusion generation process by  Semantic Diffusion Guidance (SDG) with language and/or a reference image. SDG is iteratively injected at each step of generation process. 
We only illustrate the guidance at one timestep $t$ in the figure.
}
\vspace{-5mm}
\label{fig:model}
\end{center}
\end{figure}

\subsection{Guiding Diffusion Models for Controllable Image Synthesis}\label{subsec:guiding}

Diffusion models define a Markov chain where random noise is gradually added to the data, known as the forward process. Formally, given a data point sampled from a real-data distribution $x_0 \sim q(x)$, the forward process sequentially adds Gaussian noise to the sample over $T$ timesteps:
\begin{equation}\label{eq:1}
\begin{split}
    q(x_t|x_{t-1})   &= \mathcal{N}(x_t; \sqrt{1 - \beta_{t}}x_{t-1}, \beta_{t}\mathbf{I}) \\
    q(x_{1:T}|x_{0}) &= \prod_{t=1}^{T}q(x_{t}|x_{t-1}),
\end{split}
\end{equation}
where $\{\beta\}_{t=1:T}$ denotes a constant or learned variance schedule that controls the noise step size.
A property of the forward process is that we can sample $x_t$ from $x_0$ in a closed form:
\begin{equation}\label{eq:1.2}
    q(x_t|x_0) = \sqrt{\bar{\alpha}_t}x_0+\epsilon\sqrt{1-\bar{\alpha}_t}, \epsilon \sim \mathcal{N}(0, 1)
\end{equation}
 where $\alpha_{t} = 1 - \beta_{t}$ and $\overline{\alpha_{t}} =
\prod_{s=1}^{t}\alpha_{s}$.
 
Generative modeling is done by learning the backward process where the forward process is reversed via a parameterized diagonal Guassian transition: 
\begin{equation}\label{eq:2}
p_\theta(x_{t-1}|x_{t}) = \mathcal{N}(x_{t-1}; \mu_\theta(x_{t}), \sigma^{2}_\theta(x_{t})\mathbf{I})
\end{equation}
We choose the notation $p_\theta(x_{t-1}|x_{t}) = \mathcal{N}(\mu_\theta, \sigma^{2}_\theta\mathbf{I})$ for brevity.
In order to learn the backward process, neural networks are trained to predict $\mu_\theta$ and $\sigma^{2}_\theta$. 

The formulations above explain the unconditional backward process $p_{\theta}(x_{t-1}|x_{t})$; with an extra guidance signal $y$, the sampling distribution becomes:
\begin{equation}\label{eq:5}
    p_{\theta,\phi}(x_{t-1}|x_{t},y)=Zp_\theta(x_{t-1}|x_{t})p_\phi(y|x_{t-1}),
\end{equation}
where $Z$ is a normalizing constant. It is proven~\cite{dhariwal2021diffusion} that the new distribution after incorporating the guidance can be approximated by a Gaussian distribution with shifted mean:
\begin{equation}\label{eq:6}
    p_\theta(x_{t-1}|x_{t})p_\phi(y|x_{t-1})=\mathcal{N}(\mu+\Sigma g, \Sigma),
\end{equation}
where $\mu = \mu_\theta$, $\Sigma = \sigma^{2}_\theta\mathbf{I}$, $g=\nabla_{x_{t-1}}\log p_\phi(y|x_{t-1})$.

Class-guided synthesis was explored in \cite{dhariwal2021diffusion} where $y$ is a discrete class label, and $p_\phi(y|x_{t-1})$ is the probability of $x_{t-1}$ belonging to class $y$. 
Here, we generalize $y$ to a continuous embedding for language, image or multimodal guidance. In the following, we introduce the guidance function $F_{\phi}(x_{t}, y, t)=\log p_\phi(y|x_{t})$ for different guidance types.

Figure~\ref{fig:model} and Algorithm~\ref{alg:sdg} summarize the proposed Semantic Diffusion Guidance.
Note that there is an additional scaling factor $s$ for semantic guidance in Algorithm \ref{alg:sdg}, a user-controlled hyperparameter that determines the strength of the guidance. We discuss its effect in Section \ref{sec:experiments}.

\vspace{-3mm}
\RestyleAlgo{ruled}
\begin{algorithm}
\caption{Semantic Diffusion Guidance}\label{alg:sdg}
\SetKwInput{KwData}{Input}
\KwData{guidance $y$,  scaling factor $s$}
\textbf{Given:} diffusion model ($\mu_{\theta}$,  $\sigma_{\theta}$), Guidance function $F_{\phi}(x_{t}, y, t)$  \\
$x_{T} \leftarrow$ sample from $\mathcal{N}(0, \mathbf{I})$ \\
\textbf{for} $t=T, \cdots 1$ \textbf{do} \\
    \Indp $\mu, \Sigma \leftarrow \mu_{\theta}, \sigma^{2}_{\theta}\mathbf{I}$  \\
    $x_{t-1} \leftarrow$ sample from $\mathcal{N}(\mu + s \Sigma \nabla_{x_{t}} F_{\phi}(x_{t}, y, t), \Sigma)$ \\
\Indm\textbf{end for} \\
\textbf{return} $x_0$
\end{algorithm}

\subsection{Language Guidance}\label{subsec:language}

Language is one of the most intuitive ways that a user can control the generation model. In order to incorporate language information to the image synthesis process, we use a visual-semantic embedding model for image-text alignment. Specifically, given an image $x$ and a text prompt $l$, the model embeds them into the joint embedding space using an image encoder $E_I$ and a text encoder $E_L$, respectively. The similarity between the embeddings $E_I(x)$ and $E_L(l)$ is calculated as the cosine distance, and we utilize this to formulate the language guidance function.

However, the models for backward process and guidance in Equation~\ref{eq:6} are time-dependent, and take noisy images as input. This means that the image encoder $E_I$ needs to incorporate the timestep $t$ as input and be further trained on noisy images at different timesteps as well. We denote such time-dependent image encoder for noisy images as $E^{\prime}_{I}$. Finally, the language guidance function can be defined as:
\begin{equation}
    F(x_t, l, t)=E^\prime_I(x_t, t) \cdot E_L(l),
\end{equation}
where $E^\prime$ denotes the image encoder trained on noisy images with additional timestep input. 
In Section~\ref{subsec:clip}, we give details on adapting a CLIP model~\cite{radford2021learning} to become time-dependent with minimal architecture changes, and present a self-supervised finetuning strategy for noisy images.

\subsection{Image Guidance}\label{subsec:image}

Sometimes, an image can convey information that is difficult to express in language. 
For example, users may want to generate a photo of a cat that looks similar to another cat, or want to generate a photo of a bedroom in the style of Van Gogh's painting ``The Starry Night''. 
They may also want to generate realistic images given an emoji or a  painting. 
We thus propose an approach for image-guided diffusion 
that effectively controls the content or style information according to an image.
We present two types of image guidance, namely \textit{image content guidance} and \textit{image style guidance}.

\noindent\textbf{Image Content Guidance} aims to control the content of the generated image, with or without structural constraints, based on a reference, and is formulated as the cosine similarity of the image feature embeddings. Let $x^\prime_0$ denote the noise-free reference image. We perturb $x^{\prime}_0$ per Equation \ref{eq:1.2} to get  $x^{\prime}_t$. Then, the guidance signal at timestep $t$ is,
\begin{equation}\label{eq:isg}
    F(x_t, x^\prime_t, t) = E^\prime_I(x_t, t) \cdot E^\prime_I(x^\prime_t, t).
\end{equation}

Similar to language guidance, we use an image encoder finetuned with noised images to define the image guidance function and extract embeddings that mostly capture the high-level semantics. 
An interesting property of using image encoders for guidance is that one can control how much structural information such as pose and viewpoint is maintained from the reference image. For instance, the embeddings used in Equation \ref{eq:isg} do not have spatial dimensions, resulting in samples with great variations in pose and layout. However, by utilizing spatial feature maps and forcing alignment between features in corresponding spatial locations, we can guide the generated image to additionally share similar structure with the reference image as follows.
\begin{equation}\label{eq:structure}
    F(x_t, x^\prime_t, t) =  -\sum_j\frac{1}{C_j H_j W_j}||E^{\prime}_I(x_t, t)_{j}-E^{\prime}_{I}(x^\prime_t, t)_{j}||_2^2
\end{equation}
where $E^{\prime}_{I}()_{j} \in \mathcal{R}^{C_j \times H_j \times W_j}$ denotes the spatial feature maps of the $j$-th layer of the image encoder $E^{\prime}_{I}$. 

\noindent\textbf{Image Style Guidance} allows style transfer from the reference image. It is formulated similarly, except the alignment between the Gram matrices of the intermediate feature maps is enforced: 
\begin{equation}\label{eq:style}
    F(x_t, x^\prime_t, t)=-\sum_j||G^{\prime}_I(x_t, t)_{j}-G^{\prime}_I(x^\prime_t, t)_{j}||_F^2,
\end{equation}
where $G^{\prime}_{I}()_j$ is the Gram matrix~\cite{johnson2016perceptual} of the $j$-th layer feature map of the image encoder $E^{\prime}_{I}$.

\input{sections/table_1_new}

\subsection{Multimodal Guidance}

In some application scenarios, image and language may contain complementary information, and allowing both image and language guidance at the same time provides further flexibility for user control. Our pipeline can easily incorporate both by a weighted sum of the two guidance functions, with their scaling factors as weights.
\begin{equation}
    F_{\phi_0}(x_t, y, t)=s_1 F_{\phi_1}(x_t, y, t) + s_2 F_{\phi_2}(x_t, y, t).
\end{equation}
By adjusting the weighting factors of each modality, users can balance between the language and image guidance.

\subsection{Self-supervised Finetuning of CLIP without Text Annotations}\label{subsec:clip}

CLIP~\cite{radford2021learning} is a powerful vision and language model trained on large-scale image-text data. We leverage its semantic knowledge to achieve controllable synthesis for diffusion models. To act as a guidance function, CLIP is expected to handle noisy images $x_t$ at any timestep $t$. We make a minor architectural change to CLIP image encoder $E_I$ to accept an additional input $t$ by converting batch normalization layers to adaptive batch normalization layers, where the prediction of scale and bias terms is conditioned on $t$. We denote this modified CLIP image encoder as $\widetilde{E_I}$. The parameters of $\widetilde{E_I}$ are initialized by the parameters of the pretrained CLIP model $E_I$, except for the parameters for
the adaptive batch normalization layers.

To finetune $\widetilde{E_I}$, we propose a self-supervised approach in which we force an alignment between features extracted from clean and noised images. Formally, given a batch of $N$ pairs of clean and noised images $\{x^{i}_0, x^{i}_{t_i}\}_{i=1}^{N}$ where $t_i$ is the timestep sampled for the $i$-th image that governs the amount of noise, we encode $x^{i}_0$ and $x^{i}_{t_i}$ with $E_I$ and $\widetilde{E_I}$, respectively. We rely on CLIP's contrastive objective to maximize the cosine similarity of the $N$ positive pairs while minimizing the similarity of the remaining negative pairs. We fix the parameters of $E_I$ and use the contrastive objective to finetune the parameters of $\widetilde{E_I}$.
With our finetuned CLIP model, the diffusion model can be guided by image or language information that users provide. Moreover, the CLIP model is finetuned in a self-supervised manner without requiring any language data for the target dataset.

%% file: sections/table_1_new.tex
\begin{table*}[t]
\caption{Quantitative evaluation of our proposed SDG and comparison to prior work on FFHQ dataset with image guidance and text guidance. For FID, the lower, the better. For other scores, the higher, the better.}\label{tab:comparison}
\centering
\begin{tabular}{cccccccc}
\hline
 &  & Quality & Diversity & \multicolumn{4}{c}{Correctness (retrieval evaluation)} \\ \hline
 &  & FID & LPIPS & Top 1 & Top 5 & Top 10 & Top 20 \\ \hline
\multirow{2}{*}{\begin{tabular}[c]{@{}c@{}}image guidance\end{tabular}} & ILVR (N=32)~\cite{choi2021ilvr} & 17.15 & 0.439 & 0.205 & 0.416 & 0.556 & 0.727 \\
 & SDG & \textbf{14.37} & \textbf{0.583} & \textbf{0.520} & \textbf{0.742} & \textbf{0.816} & \textbf{0.906} \\ \hline
\multirow{2}{*}{\begin{tabular}[c]{@{}c@{}}text guidance\end{tabular}} 
 & StyleGAN+CLIP & 57.45 & 0.578 & \textbf{0.749} & \textbf{0.934} & \textbf{0.974} & \textbf{0.996} \\
 & SDG & \textbf{28.38} & \textbf{0.610} & 0.553 & 0.795 & 0.878 & 0.947 \\ \hline
\end{tabular}
\end{table*}

%% file: sections/experiments.tex
\section{Experiments}\label{sec:experiments}

\input{sections/table_2}
\input{sections/table_3}

\begin{figure*}[t]
\begin{center}
\vspace{-4mm}
\includegraphics[width=0.95\linewidth]{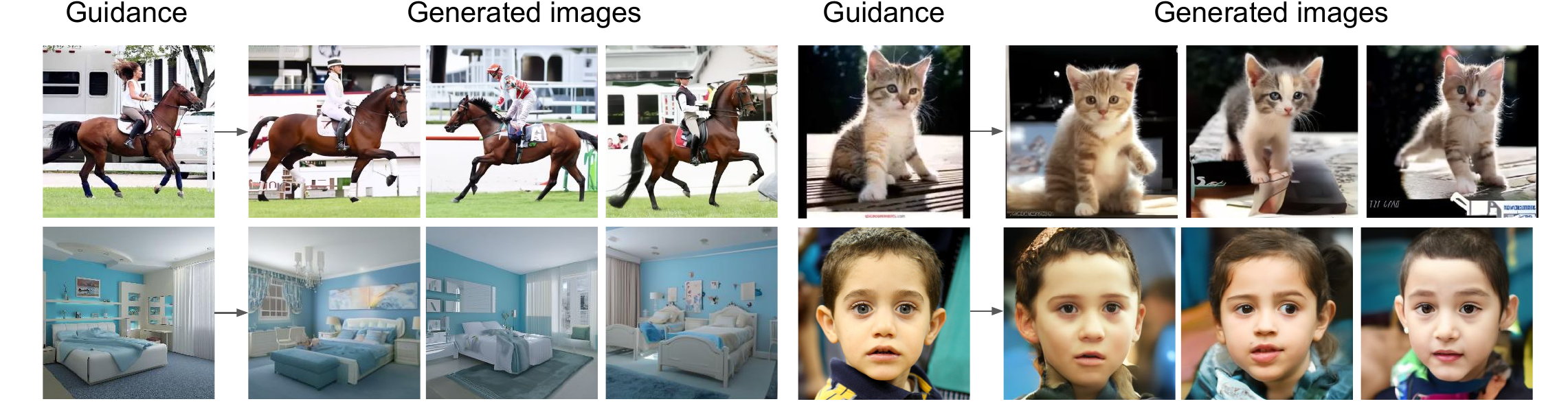}
\vspace{-3mm}
\caption{Image synthesis results with image content guidance on LSUN and FFHQ datasets. Given a guidance image, the model is able to generate semantically similar images with different pose, layout, and structure.}
\vspace{-5mm}
\label{fig:result_v}
\end{center}
\end{figure*}

\begin{figure*}[t]
\begin{center}
\includegraphics[width=0.95\linewidth]{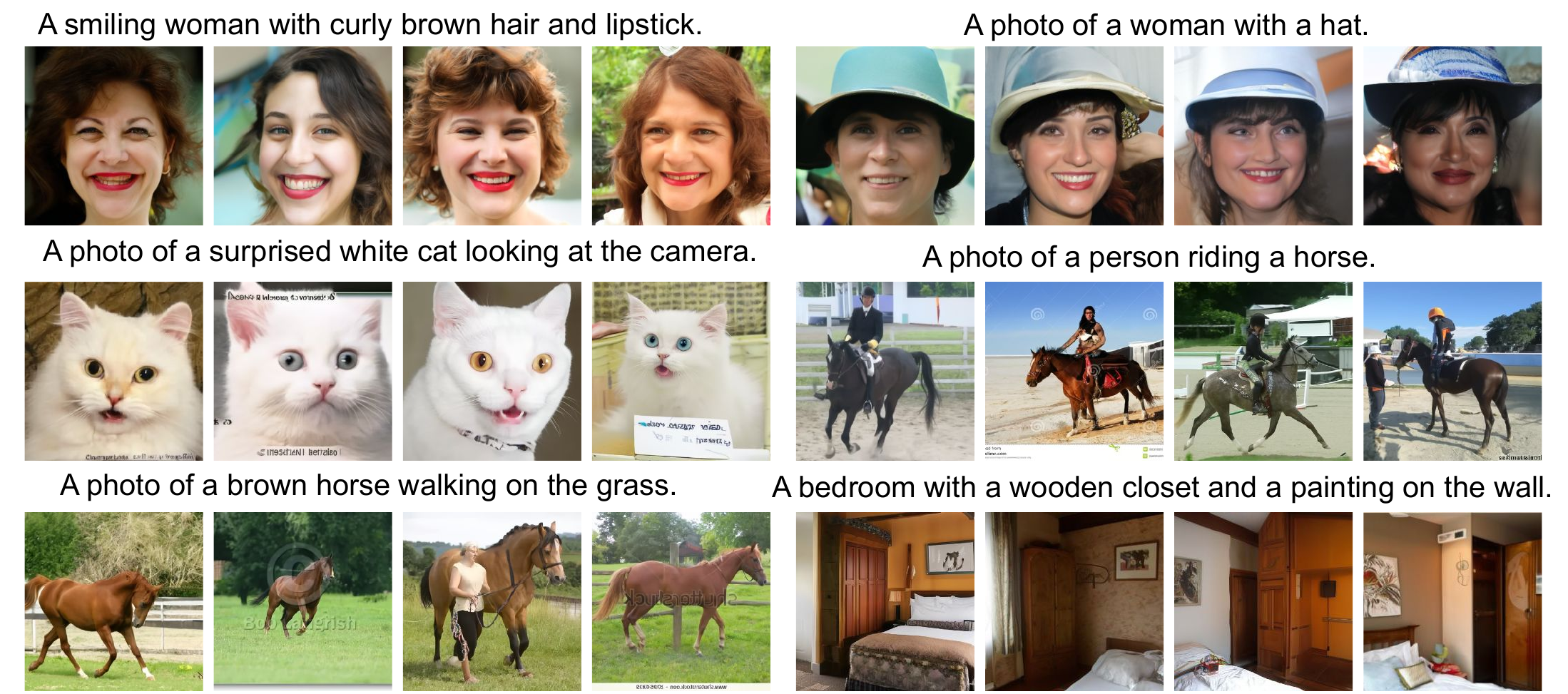}
\vspace{-3mm}
\caption{Image synthesis results with language guidance on LSUN and FFHQ datasets. Our model is able to generate images based on fine-grained language instructions.}
\vspace{-5mm}
\label{fig:result_l}
\end{center}
\vspace{-2mm}
\end{figure*}

\begin{figure*}[t]
\begin{center}
\includegraphics[width=0.95\linewidth]{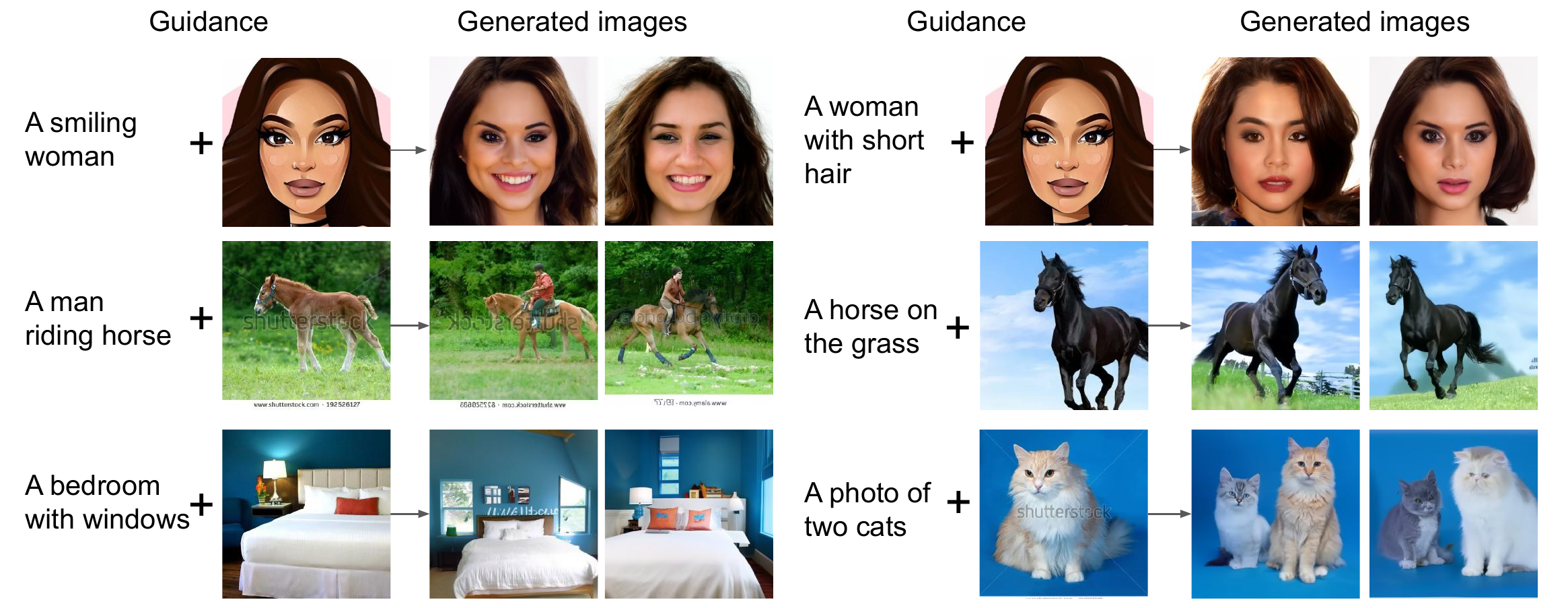}
\vspace{-3mm}
\caption{Image synthesis results with both image and language guidance. The image and language guidance provides complementary information, and our model generates images that matches both sources of guidance.}
\vspace{-5mm}
\label{fig:result_vl}
\end{center}
\end{figure*}

\begin{figure*}[t]
\begin{center}
\vspace{-3mm}
\includegraphics[width=0.95\linewidth]{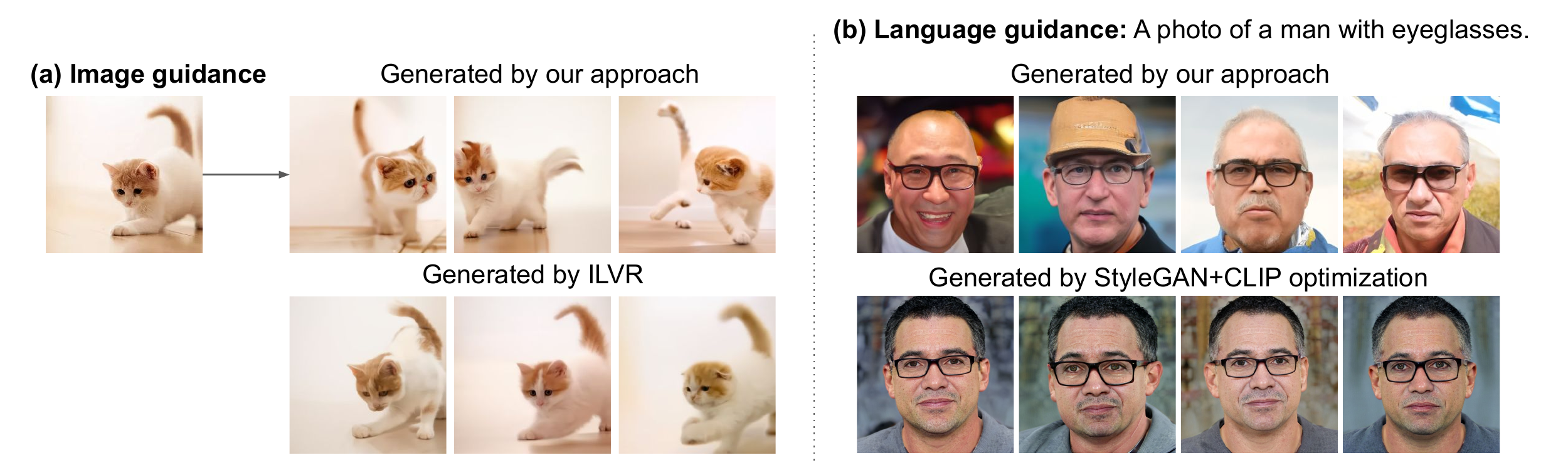}
\vspace{-3mm}
\caption{Comparison to previous work. (a) Image-guided image synthesis is compared with ILVR, (b) text-guided image synthesis is compared with StyleGAN+CLIP.}
\vspace{-3mm}
\label{fig:result_compare}
\vspace{3mm}
\includegraphics[width=0.95\linewidth]{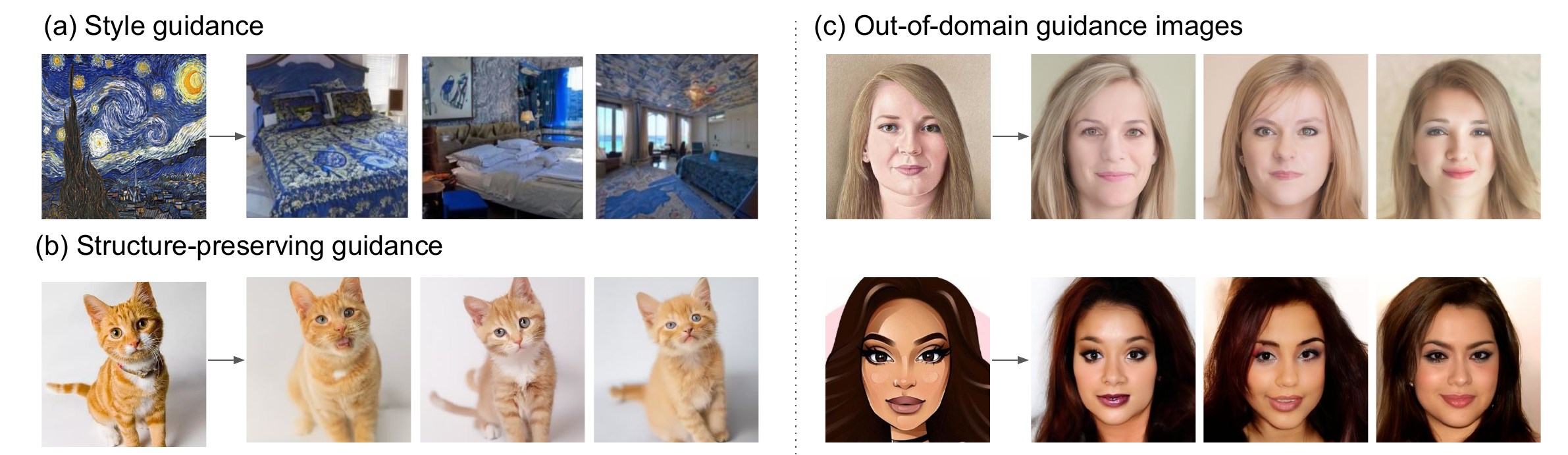}
\vspace{-3mm}
\caption{Different applications of SDG. (a) Style-guided synthesis. (b) Structure-preserving synthesis when the user does not want to generate diverse structures. (c) Synthesizing photo-realistic images with out-of-domain image guidance.}
\vspace{-5mm}
\label{fig:result_misc}
\end{center}
\end{figure*}

\subsection{Dataset and Implementation Details}

We conduct experiments on FFHQ~\cite{karras2019style} and LSUN~\cite{yu2015lsun} cat, horse, and bedroom subsets. FFHQ dataset contains 70,000 images of human faces. LSUN contains 3 million bedroom images, 2 million horse images, and 1.7 million cat images. We use unconditional DDPMs from~\cite{dhariwal2021diffusion,choi2021ilvr}, and finetune CLIP~\cite{radford2021learning} RestNet 50$\times$16 models on noised images on each dataset with initial learning rate $10^{-4}$ and weight decay $10^{-3}$, with a batch size of 256. 
When synthesizing images with our SDG, the scaling factor is a hyperparameter that we manually adjust for each guidance, which will be discussed in Sec.~\ref{subsec:ablation}. The default scaling factor is 100 for image guidance and 120 for language guidance. 

\subsection{Quantitative Evaluation}

\noindent\textbf{Evaluation Setup}
Since our SDG is the first method that unifies text guidance and image guidance for image synthesis, there is no previous work on image synthesis with both image and language guidance. So we evaluate the language-guided image synthesis and image-guided image synthesis separately in order to compare with previous work.
We evaluate the \textbf{language-guided} generation on FFHQ dataset. For that we define 400 text instructions based on combinations of gender and face attributes from CelebA-Attributes~\cite{liu2015faceattributes}. For example, ``A photo of a smiling man with glasses''.
We generate 25 images for each text query, which results in 10,000 images in total. 
We compare our language-guided generation with StyleGAN+CLIP\footnote{\url{https://colab.research.google.com/drive/1br7GP\_D6XCgulxPTAFhwGaV-ijFe084X}}, which uses CLIP~\cite{radford2021learning} loss to optimize the randomly initialized latent codes of StyleGAN~\cite{karras2019style} for text-guided image synthesis. StyleGAN+CLIP removes the GAN inversion module of StyleCLIP~\cite{karras2019style} so that it can be applied for language-based image synthesis. Since our model does not require text annotation for training, our text-guided image synthesis experiments are conducted on image-only datasets without paired text annotations. So our method cannot be directly compared with other text-based image synthesis methods which have to be trained on text-image paired datasets.
To evaluate \textbf{image-guided} image synthesis, we randomly choose 10,000 images from each dataset as guidance and synthesize new images based on the guidance images. 
We compare our image-guided results to ILVR~\cite{choi2021ilvr}.

We present quantitative results and comparison with previous work in Table~\ref{tab:comparison} with the following evaluation metrics.

\noindent\textbf{FID for image quality evaluation.} We report FID score~\cite{heusel2017gans} calculated on 10,000 images
for each dataset to evaluate the quality of generated images. Lower FID indicates better generation quality. Our SDG outperforms comapred methods for both image-guided synthesis and language-guided synthesis.

\noindent\textbf{LPIPS for diversity evaluation.} We calculate the LPIPS score~\cite{zhang2018perceptual} between paired images generated from the same image guidance or the same text guidance, as shown in Table~\ref{tab:comparison}. Higher LPIPS indicates more diversity. Our model generates more diverse images compared to previous work ILVR~\cite{choi2021ilvr} and StyleGAN+CLIP. 
The images generated by ILVR follows the same structure and layout, with variations in details. While our method is able to generate diverse images with different pose, structure, and layout, as shown in Figure~\ref{fig:result_compare}(a). The images generated by StyleGAN+CLIP also suffers from low diversity, as shown in Figure~\ref{fig:result_compare}(b). The high FID score of StyleGAN+CLIP is also because of the low diversity of the generated images.

\noindent\textbf{Retrieval accuracy to evaluate consistency with guidance.} We use text-to-image retrieval or image retrieval by an original CLIP ResNet 50$\times$16 model without finetuning to evaluate how well the generated images matches the guidance. For an image generated with text guidance, we randomly select 99 real images from the training set as negative images, and evaluate the text-to-image retrieval performance. Similarly, for an image synthesized with a reference image, we use the reference image to retrieve the generated image from the 99 randomly selected real images\footnote{The selected negative images are disjoint with the guidance images we used for synthesizing images.}. StyleGAN+CLIP has a very high retrieval performance because the latent codes of the StyleGAN model are directly optimized to minimize the CLIP score calculated by the CLIP model used for retrieval. So the high retrieval performance of StyleGAN+CLIP comes at the cost of low generation diversity, as indicated by the high FID and low LPIPS scores.

\subsection{Ablation Study}\label{subsec:ablation}
As demonstrated in Section~\ref{subsec:guiding} and Algorithm~\ref{alg:sdg}, the scaling factor $s$ is a user-controllable hyper-parameter that controls the strength of the guidance. We explore the effect of the scaling factor in Table~\ref{tab:ablation_image} and Table~\ref{tab:ablation_text}. 
A visual example of the effect of different scaling factors is shown in the Appendix.
We observe the trade-off between semantic correctness and diversity of generated images. As the scaling factor gets larger, the guidance signal has more control on the generation results, as indicated by the increased semantic consistency with the guidance. While larger scaling factor also leads to lower diversity of generated images. Users can adjust the scaling factor to control how diverse they expect the generated images to be.

\subsection{Qualitative Results}

\noindent\textbf{Text-guided and image-guided synthesis results} Our model combines the language and image guidance in a unified framework, and is easy to adapt to various applications. In Figure~\ref{fig:result_v} we show the synthesis results with image content guidance (Equation~\ref{eq:isg}). With the image guided diffusion, the model is able to synthesize new images with diverse structures that match the semantics of the guidance image. Figure~\ref{fig:result_l} shows the language-guided diffusion results, where our model is able to handle complex and fine-grained descriptions, such as ``A smiling woman with curly brown hair and lipstick.'', or ``A bedroom with a wooden closet and a painting on the wall.'' 
We can also incorporate language and image guidance jointly, as shown in Figure~\ref{fig:result_vl}. The image and language guidance provide complementary information, and our semantic diffusion guidance is able to generate images that align with both. For example, we can generate a bedroom similar to the guidance bedroom image but with windows, or generate a woman according to a guidance image but with a new attribute defined the language guidance (e.g., ``smiling'' or ``short hair'' or ``sunglasses'').

\noindent\textbf{Comparison to prior work} Since there is no prior work that incorporates text and image guidance in the same unified framework, we compare our approach to previous text-guided and image-guided synthesis work. In image-guided synthesis, the most related to our work is ILVR~\cite{choi2021ilvr}.
As shown in Fig.~\ref{fig:result_compare}(a), our model can generate images in different poses and structures, while ILVR can only generate images of the same pose and structure. We compare our language-guided image synthesis with StyleGAN+CLIP in Fig.~\ref{fig:result_compare}(b). Although StyleGAN+CLIP is able to generate high-quality images, diversity is lacking in their results, while our model is able to generate high-quality and diverse results based on the language instructions. 

\noindent\textbf{Other applications} In Fig.~\ref{fig:result_misc}(a,b), we demonstrate the results of style (Equation~\ref{eq:style}) and structure-preserving (Equation~\ref{eq:structure}) image guidance. With the style guidance, the model trained on LSUN bedroom is able to synthesize bedrooms in the unseen style. With the structure-preserving content guidance, the synthesized images preserve the structure, pose, and layout from the reference image. Fig.~\ref{fig:result_misc}(c) shows that the model is able to take an out-of-domain image as guidance, and synthesize photo-realistic images which are semantically similar to the guidance cartoon image.

%% file: sections/table_2.tex
\begin{table*}[t]
\caption{Ablation study of our proposed SDG with image guidance. The numbers in the brackets after ``SDG'' indicates the scaling factor. For FID, the lower, the better. For other scores, the higher, the better.}\label{tab:ablation_image}
\vspace{-3mm}
\centering
\begin{tabular}{cccccccc}
\hline
 &  & Quality & Diversity & \multicolumn{4}{c}{Correctness (retrieval evaluation)} \\ \hline
 &  & FID & LPIPS & Top 1 & Top 5 & Top 10 & Top 20 \\ \hline
\multirow{2}{*}{\begin{tabular}[c]{@{}c@{}}LSUN\\ Cat\end{tabular}}
 & SDG (100) & \textbf{16.02} & \textbf{0.617} & 0.178 & 0.443 & 0.592 & 0.766 \\
 & SDG (200) & 16.23 & 0.565 & \textbf{0.278} & \textbf{0.533} & \textbf{0.738} & \textbf{0.880} \\ \hline
\multirow{2}{*}{\begin{tabular}[c]{@{}c@{}}LSUN\\ Horse\end{tabular}}
 & SDG (100) & \textbf{10.30} & \textbf{0.597} & 0.165 & 0.418 & 0.568 & 0.704 \\
 & SDG (200) & 11.22 & 0.585 & \textbf{0.298} & \textbf{0.609} & \textbf{0.738} & \textbf{0.863} \\ \hline
\multirow{2}{*}{\begin{tabular}[c]{@{}c@{}}LSUN\\ Bedroom\end{tabular}}
 & SDG (100) & \textbf{5.18} & \textbf{0.633} & 0.364 & 0.745 & 0.866 & 0.942 \\
 & SDG (200) & 5.19 & 0.550 & \textbf{0.445} & \textbf{0.805} & \textbf{0.900} & \textbf{0.951} \\ \hline
\end{tabular}
\end{table*}

%% file: sections/table_3.tex
\begin{table*}[t]
\vspace{-3mm}
\caption{Ablation study of our proposed SDG with language guidance on FFHQ dataset. The numbers in the brackets after ``SDG'' is the scaling factor. For FID, the lower, the better. For other metrics, the higher, the better.}\label{tab:ablation_text}
\vspace{-3mm}
\centering
\begin{tabular}{cccccccc}
\hline
 &  & Quality & Diversity & \multicolumn{3}{c}{Correctness (retrieval accuracy)} \\ \hline
 &  & FID & LPIPS & Top 1 & Top 5 & Top 10 & Top 20 \\ \hline
\multirow{4}{*}{FFHQ} 
 & SDG (120) & \textbf{19.60} & \textbf{0.650} & 0.248 & 0.526 & 0.654 & 0.795 \\
 & SDG (160) & 22.63 & 0.644 & 0.263 & 0.548 & 0.679 & 0.801 \\
 & SDG (320) & 28.38 & 0.610 & \textbf{0.553} & \textbf{0.795} & \textbf{0.878} & \textbf{0.947} \\ \hline
\end{tabular}
\vspace{-3mm}
\end{table*}
\vspace{-3mm}

%% file: sections/conclusion.tex
\section{Conclusion and Discussions}\label{sec:conclusion}
We propose Semantic Diffusion Guidance (SDG), a unified framework for diffusion-based image synthesis with language, image, or multi-modal guidance. 
The flexible guidance module allows us to inject various types of guidance into any off-the-shelf unconditional diffusion model without re-training or finetuning the diffusion model.
We further present a self-supervised efficient finetuning scheme for the CLIP guidance model which does not require textual annotations.
However, image generation has as much potential for misuse as it has for beneficial applications. We should be aware of the potential negative social impact if image synthesis is used for generating fake images to mislead people.

\section*{Acknowledgements}
This work was supported in part by DoD including DARPA’s SemaFor, PTG and/or LwLL programs, as well as BAIR’s industrial alliance programs.